Dec 2nd, 9:00 AM - Dec 5th, 5:00 PM

# Synthetic Realities: Evaluating Human Ability to Distinguish AI-Generated Videos from Real Footage

Danyal Sarfraz
*Design Department, Ulsan National Institute of Science and Technology, Ulsan, South Korea*

Raja Mubashar Karim
*School of Art Design & Architecture, National University of Science and Technology, Islamabad, Pakistan*

KwanMyung Kim
*Design Department, Ulsan National Institute of Science and Technology, Ulsan, South Korea*

# Synthetic Realities: Evaluating Human Ability to Distinguish AI-Generated Videos from Real Footage

Sarfraz, Danyal[a]; Mubashar Karim, Raja[b]; Kim, KwanMyung[a]*

[a] Design Department, Ulsan National Institute of Science & Technology (UNIST), Ulsan, South Korea
[b] School of Art Design & Architecture (SADA), National University of Science & Technology, Islamabad, Pakistan
* kmyung@unist.ac.kr

Recent advances in generative AI have enabled multiple pathways for high-fidelity video synthesis; text-to-video, image-to-video animation, and video outpainting. Empirical side-by-side evaluations of how untrained viewers perceive and distinguish these outputs from real footage remain scarce. In this study, we systematically compare human detection accuracy across these three AI generation techniques within three thematic contexts: historical footage, film and media content, and natural environments. We constructed a balanced stimulus set comprising equal amounts of real and AI-generated videos (18 each). The AI clips were evenly distributed across the three generation methods using Google's Veo 3, Lightricks LTX Video, and Wan VACE. All videos were produced and standardized within the ComfyUI framework to ensure consistent quality and duration. 87 participants judged every clip in a binary forced-choice task ("real" vs. "AI-generated"). Participants correctly identified videos 60% of the time on average. Image-to-Video clips were recognized most accurately (79%), followed by real footage (64%), outpainting (49%), and text-to-video (43%). Accuracy also varied by theme: film and historical scenes yielded higher detection rates than environmental clips, which were frequently mistaken for AI. Logistic regression confirmed significant effects of both technique and theme as well as their interaction ($p < 0.001$), indicating that detection success depends jointly on how the content was generated and what it depicts. Findings reveal a consistent bias toward assuming synthetic origins and highlight that perceptual realism in AI video is shaped more by context than by model type, underscoring the importance of media-literacy approaches and context-aware evaluation tools for navigating increasingly synthetic visual media.

## 1 Introduction

Recent advances in AI have produced a surge of highly realistic synthetic video content through diverse generation techniques. Modern diffusion and transformer-based models now enable multiple pathways to video synthesis: text-to-video systems that create fully scripted clips from natural language prompts, image-to-video pipelines that animate static images into dynamic sequences, and video outpainting methods that extend existing footage beyond original frame boundaries. These capabilities are rapidly being adopted by media artists and designers, expanding creative possibilities and challenging traditional notions of visual authenticity in art and design contexts.

This emerging landscape of AI-generated video techniques presents broader implications for creative practice and visual literacy. Each generation method produces distinct visual characteristics that may influence human perception differently. However, empirical research comparing these techniques side-by-side is lacking.

This study systematically compares human perception of AI-generated videos across three distinct generation techniques: text-to-video synthesis, image-to-video animation, and video outpainting, within varied thematic contexts including historical reenactments, cinematic-style scenes, and natural environments. We examine these technologies through an arts-and-design lens, recognizing their primary role as creative tools that blur the line between authentic and synthetic media.

Our research investigates two key questions: To what accuracy can different AI video generation techniques be detected by untrained viewers? And how does thematic context influence detection performance across these techniques? By evaluating detection accuracy across multiple generation methods and content domains, we can provide insights for creators, educators, and audiences and navigate the evolving landscape of AI in creative practice.

# 2 Related Work

## 2.1 Evolution of Generative Video Techniques

The past five years have witnessed a profound transformation in video synthesis, as research shifted from compartmentalized pipelines to unified end-to-end architectures. Early methods leveraged cascaded diffusion and super-resolution stages, where a text-to-image model first generated key frames and subsequent networks interpolated intermediate frames. Google's Imagen Video integrates a frozen T5 text encoder with interleaved spatial-temporal diffusion and super-resolution modules to produce high-definition sequences at 24 FPS directly from text prompts, achieving both high fidelity and controllability through progressive distillation and classifier-free guidance (Ho et al., 2022). Meta Make-A-Video leverages paired text-image data alongside unlabelled video footage that can accelerate training and yield diverse, temporally coherent clips, without requiring text-video datasets (Singer et al., 2022). Autoregressive-over-autoregressive frameworks such as NUWA-Infinity decompose videos into global patches and local tokens, employing a "render-and-optimize" strategy that models long-range dependencies for arbitrarily long, high-resolution outputs with efficient convergence (Wu et al., 2022). Transformer-augmented diffusion systems like CogVideoX further incorporate expert transformer modules to capture long-range frame dependencies, yielding smoother motion trajectories and improved temporal coherence across multi-second clips (Yang et al., 2025). More recently, Google's Veo 3 unified advanced prompt understanding, physics-based simulation, and synchronized audio generation into a single commercial pipeline. Veo 3 supports native ambient sound and lip-synced dialogue, camera control primitives for cinematic movements, and embeds SynthID watermarks for responsible deployment, marking a significant step toward fully integrated synthetic video creation.

## 2.2 Advances in Text-to-Video, Image-to-Video, and Outpainting

Generative video workflows coalesce around three primary modalities, each offering distinct advantages. Text-to-video synthesis offers factorized pipelines that decouple content planning from motion rendering. EMUVideo's two-stage approach first generates a high-quality key frame via a pretrained text-to-image model, then extends this into a full clip through a unified video diffusion network conditioned on both prompt and image (Aldausari et al., 2023). It reports superior prompt fidelity and motion realism compared to prior cascaded systems, producing clips at 512×512 resolution and 16FPS that designers prefer for narrative coherence. Lightricks' LTX-Video introduces a holistic latent-diffusion architecture wherein a Video-VAE compresses spatiotemporal data by 1:192 before a transformer performs full self-attention in latent space (HaCohen et al., 2024a). This design enables real-time generation, five seconds of 768×512 video in two seconds on an NVIDIA H100, while preserving fine details through a denoising decoder that sidesteps separate up-sampling modules.

Image-to-video pipelines extend static visuals into dynamic sequences by modelling motion trajectories within latent representations. Stable Video Diffusion fine-tunes latent diffusion on paired image-video data to generate coherent 2 to 4 second clips at up to 1024×576 resolution, preserving object identity and style across frames (Wang et al., 2024). Dreamix enhances this with optical flow and semantic guidance, granting granular control over camera and object movements while maintaining texture realism (Molad et al., 2023).

Video outpainting works by extrapolating footage beyond its original framing. Alibaba's Wan2.1-VACE consolidates these methods: text-to-video synthesis, reference-to-video, pose transfer, selective region editing, and "Expand-Anything" outpainting within a Context Adapter architecture, enabling creators to compose multi-task video edits in a single interface (Jiang et al., 2025).

## 2.3 Perceptual and Creative Applications

How different frameworks in generative video are used has been explored in various studies. Work by Barrington, Cooper, and Farid specifically reports that human participants cannot consistently identify AI-generated voices, perceiving the identity of a fake voice as the real one approximately 80% of the time (Barrington et al., 2025). A large body of literature also exists on the threats and weaponization of deepfakes, which suggests the apprehension regarding identifying AI content (Farid, 2021; Khichi & Kumar Yadav, 2021; Corcoran et al., 2021; Appel & Prietzel, 2022). The current literature may underestimate people's susceptibility to deepfakes because studies often examine detection in ideal conditions. Another related study cited in the results found that detection accuracy was less than 26% on high-quality deepfakes (Somoray et al., 2025).

On the other end, case studies in museums and educational settings illustrate the engagement potential of synthetic video when framed as creative augmentation rather than deception. AI-generated historical reenactments encourage visitors to reflect on authenticity and authorship, while animated instructor avatars have boosted student engagement by blending real and synthetic presentation styles. Studies of Runway's Gen-2 platform demonstrate that multimodal prompts-combining text and reference images-enhance user agency, with designers reporting greater satisfaction in creative control and stylistic variety under dual-conditioning workflows (Anantrasirichai et al., 2025). However, systematic side-by-side comparisons of multiple generation techniques across thematic domains: historical reenactments, cinematic scenes and environments can generate generalizable insights for applications of AI beyond deepfakes.

## 2.4 Identified Gaps and Future Directions

In contrast to the vast amount of research on deepfakes, there is a relative lack of systematic comparisons across other application domains like film, tv, media, social media footage etc. Additionally, different methods such as text-to-video, image-to-video, and outpainting have not been compared under unified experimental frameworks, leaving open questions about which techniques are most convincing or prone to artifacts in specific contexts.

Thematic context profoundly shapes perception: political deepfake studies reveal a liar's dividend effect wherein awareness of manipulation induces undue skepticism toward real content, yet analogous experiments in artistic or environmental settings have not been conducted (Somoray et al., 2025). Lastly, human limitations in detection warrant deeper investigation. Meta-analyses show that untrained viewers are confused by good-quality deepfakes in 75.5% of cases, even with incentives or warnings, and often overestimate their detection skills due to overconfidence and framing biases (Korshunov & Marcel, 2020). Temporal artifacts like motion blur, inconsistent rendering cues are fleeting and harder to spot than static anomalies, while attentional fatigue and top-down expectations further degrade accuracy.

This paper aims to create multi-technique stimulus sets spanning diverse themes, coupled with multi-metric human evaluations, accuracy, and confidence, to establish robust benchmarks.

# 3 Methodology

The study's methodology involves two key phases. First, we generated stimulus videos by applying a range of AI-based techniques and collecting real footage across different thematic contexts. Second, we conducted a perception survey to gather quantitative data on how these varying techniques and themes influence viewer judgments. A detailed depiction of this workflow is provided in Figure 1.

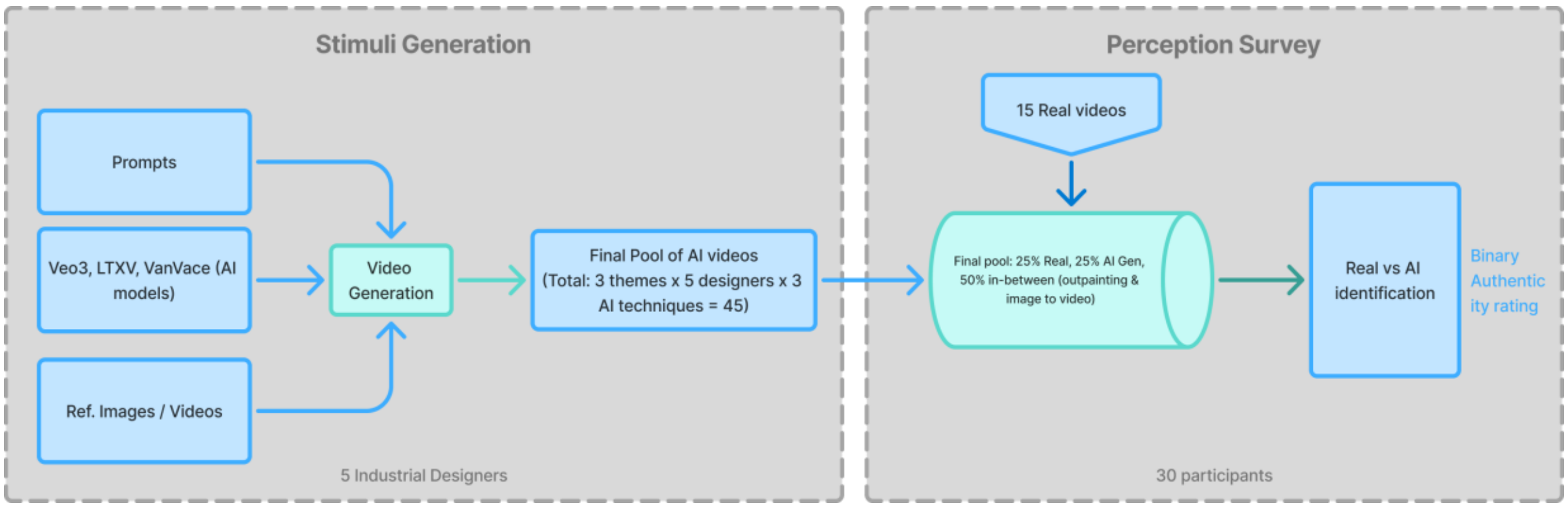

*Figure 1 Overall Workflow of the method*

## 3.1 Stimuli Generation

The stimulus set comprised a pool of videos developed by five designers employing diverse AI-based generation techniques across multiple thematic contexts. Three primary methods were utilized to produce the final video set: outpainting, image-to-video, and text-to-video generation. Outpainting expands the visible boundaries of an existing video to reveal contextual regions beyond the original frame. Image-to-video involves temporally extending a static image into a coherent video sequence. Text-to-video, in contrast, synthesizes video content directly from textual descriptions. These methods were selected to represent varying degrees of AI dependence on initial input material, ranging from high input dependency in outpainting (where a real video serves as the central reference) and image-to-video (which begins from a given image), to minimal dependency in text-to-video generation, where the entire sequence is generated from a textual prompt alone.

The video themes were also set to avoid comparison of overly heterogeneous content as pilot observations indicated AI creates better generation on few things compared like video with static objects compared to video with people or moving objects. To avoid limiting the study to one category, three broad themes were selected: environment, historical footage, and film media. The rationale was to later on also have a one v one comparison in each theme between different techniques in the same theme.

For text to video generation Google's Veo 3 was used. Veo 3 is a state-of-the-art text-to-video model capable of generating realistic, fully scripted clips from natural language prompts. Veo 3 integrates cinematic composition and physics-aware motion, with a reported "cinematic language understanding" that delivers excellent composition and movement quality. The second approach employed the Lightricks LTX Video model (LTXV), a transformer-based latent diffusion video generator designed for high frame-rate continuity (HaCohen et al., 2024). LTXV supports both text-to-video and image-to-video inputs and uses a multiscale rendering pipeline that "progressively refines" video frames to preserve motion and temporal coherence. The third method was Wan VACE, an open-source multi-task video editing model from Alibaba. VACE specializes in video outpainting or extension, which refers to completing actions or extending backgrounds beyond the original frame, thereby adding spatial context and overall scene coherence (Jiang et al., 2025). In sum, Veo 3 provided fully text-driven cinematic sequences, LTXV and Wan VACE yielded smooth interpolated motion and realistic dynamics, and Wan VACE enabled spatial expansion and consistent action continuity across frames (Figure 2).

AI clips were created using the ComfyUI workflow environment, which ensured consistent control of diffusion steps, guidance scales, and prompt weights. ComfyUI's node-based interface allowed direct replication of settings across methods, isolating the effects of generation technique rather than interface variability. Each designer generated multiple candidate clips per condition, from which the highest-quality example was selected. The review prioritized objective criteria: visual fidelity (sharpness, realism of textures), temporal coherence (smooth motion continuity), and thematic relevance (clarity of the subject and scene). Figure 3 shows an example ComfyUI workflow node graph for Wan VACE Video text-to-video generation. All models were run in such pipelines to guarantee consistency. The final stimulus set thus comprised 36 high-quality clips, evenly distributed across the three themes and three generation techniques for use in the subsequent survey.

Real clips were sourced from publicly available stock and archival footage that matched the same thematic categories and approximate duration of 8-12 seconds as the AI clips. All clips were standardized for resolution (1080p) and visual quality to eliminate low-level cues.

### 3.2 Perception Survey

The final selection of stimulus set for perception survey consisted of an equal number of real and AI-generated video clips, designed to enable direct comparison of authenticity judgments across methods and contexts. A total of 36 clips were used: 18 real videos and 18 AI-generated videos. The AI subset was evenly divided among three techniques (text-to-video, image-to-video, outpainting), each represented within the three thematic contexts of film, history, and environment. Each AI technique contributed six clips, representing two videos per thematic category (film, history, and environment).

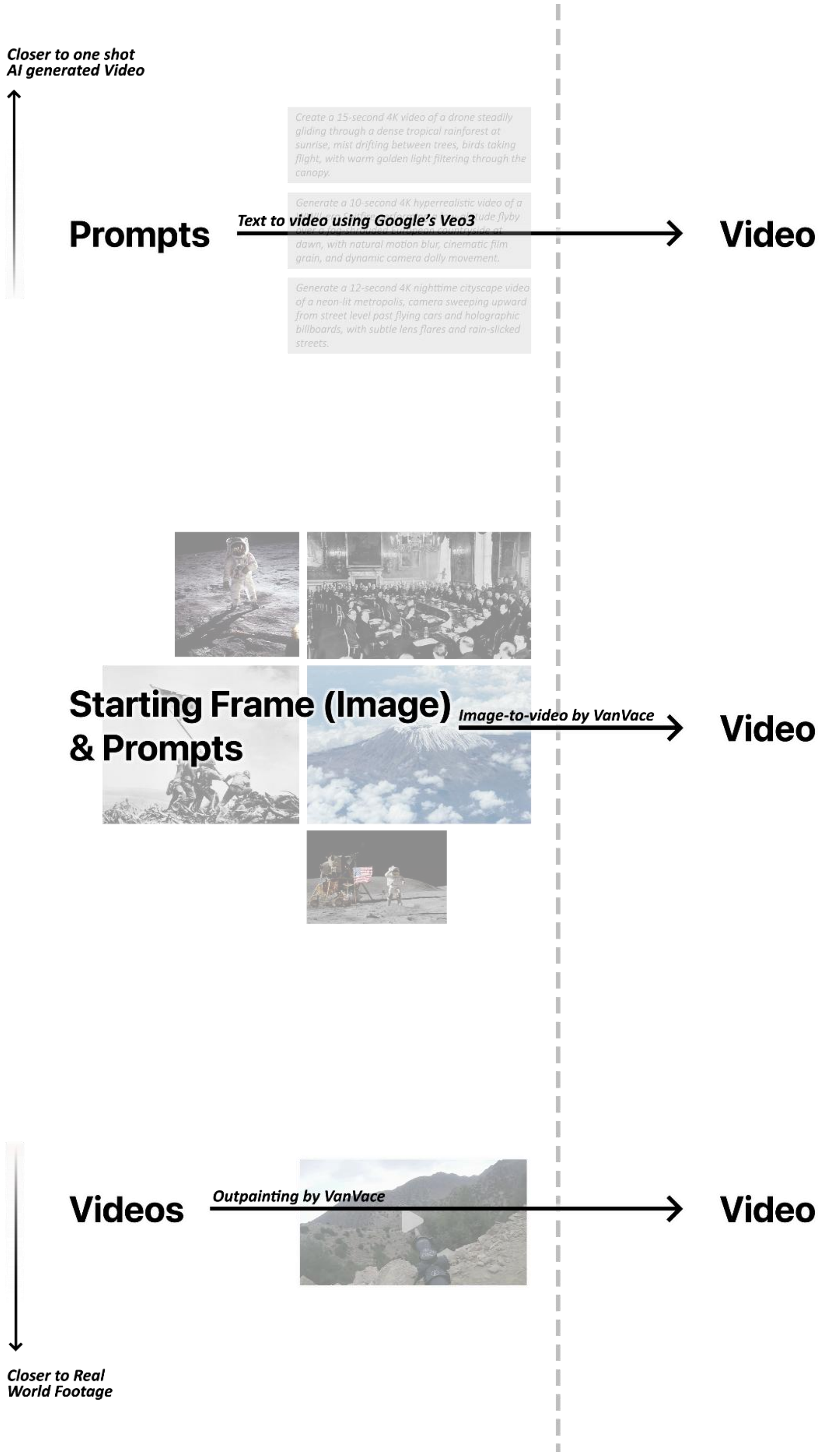

*Figure 2. Overview of the 3 types of content generated by the help of AI*

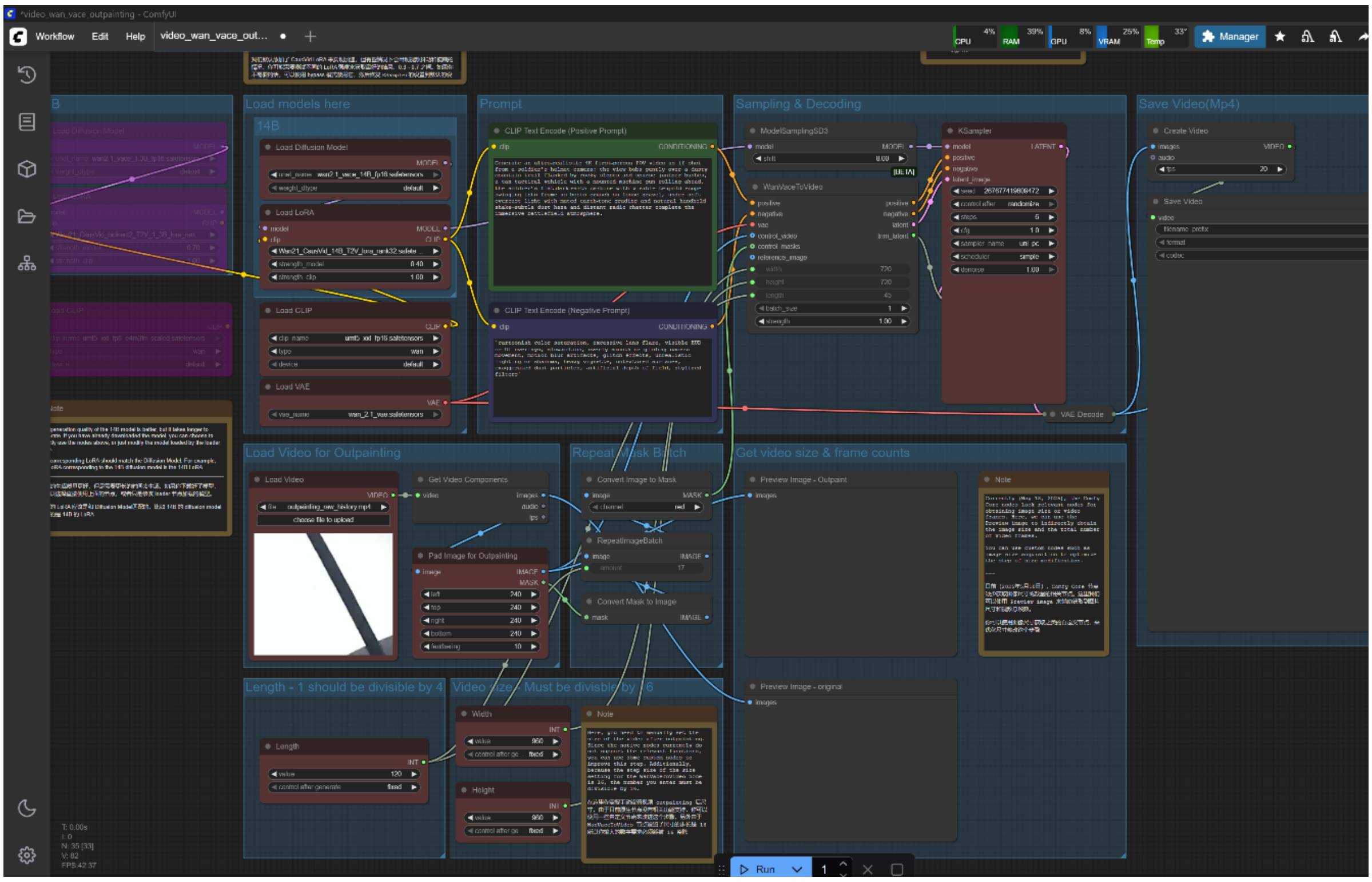

*Figure 3. ComfyUI workflow for creating WanVace based outpainting.*

A controlled online study was conducted to assess how accurately viewers could differentiate AI-generated videos from real footage. The study followed a within-subjects design in which all participants evaluated the same set of stimuli under identical conditions. 87 participants (ages 18–70) were recruited through university mailing lists and online forums. None reported professional experience with artificial intelligence, computer graphics, or video production, ensuring that their judgments reflected general-audience perception rather than domain expertise. The participants weren't informed regarding the amount of AI videos, they were just requested to identify a video is AI or real.

Each participant was presented with the complete set of 36 video clips (18 real and 18 AI-generated). The clips were displayed individually in a randomized order to eliminate sequence effects. Each video and was shown without audio to restrict evaluations to visual information only. After viewing each clip, participants answered a single question presented on-screen: "Do you believe this video is real or AI-generated?" Responses were recorded on a binary scale ("Real" / "AI-generated"). Participants could replay each clip as many times before submitting a response. To minimize fatigue, participants were encouraged to take short breaks after 18 clips. Instructions emphasized reliance on visual cues such as motion continuity, edge artifacts, and rendering consistency.

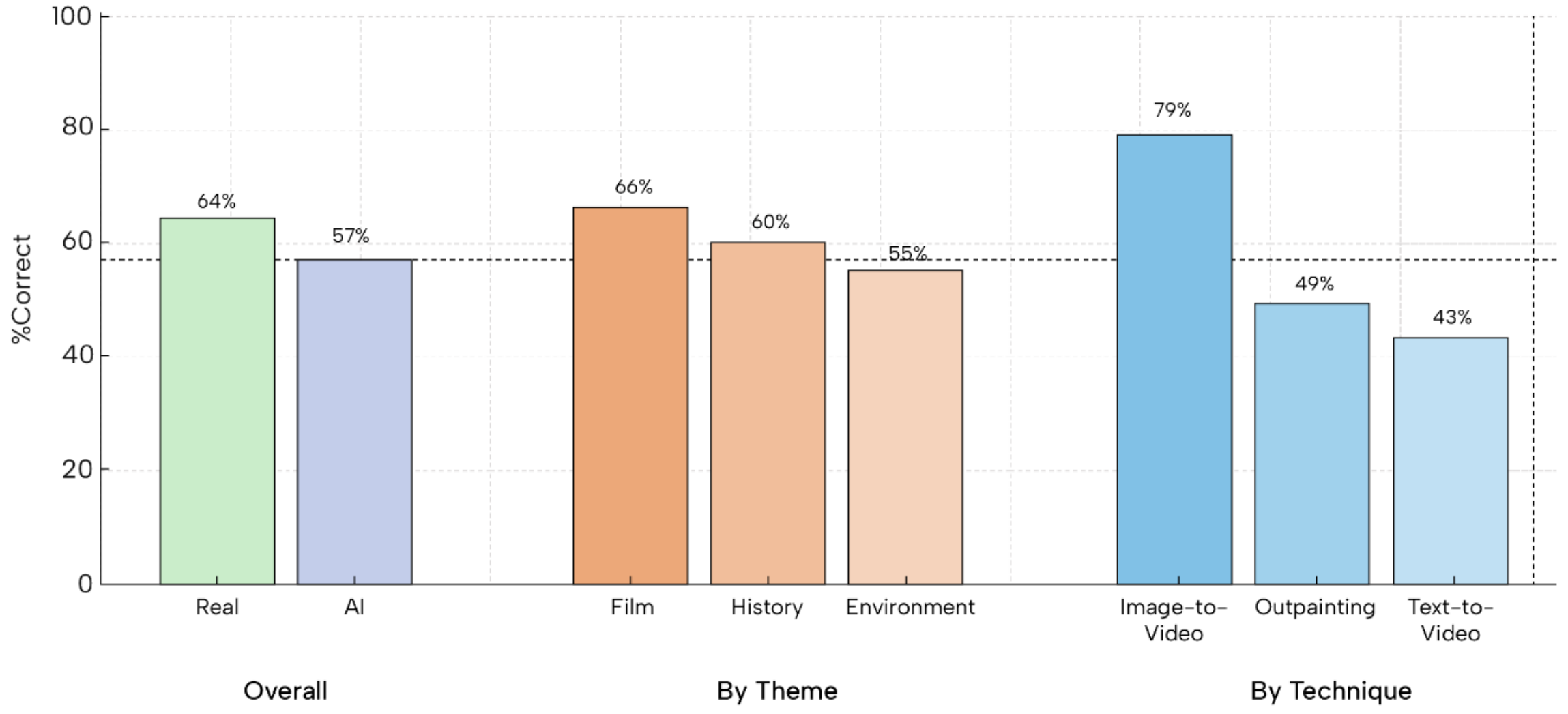

*Figure 4. Results of overall research: Real vs AI, furthermore, results compared with in techniques and themes.*

# 4 Results

Each participant's judgment was coded as a binary outcome (1 = correct, 0 = incorrect). For real clips, a "Real" response was coded as correct; for AI-generated clips, an "AI" response was coded as correct. The dataset comprised 87 participants and 36 stimuli (18 real and 18 AI-generated), producing 3,132 total observations.

To explore how accurately participants could identify real and AI videos, the data were analysed using a logistic regression model. This type of analysis is suited for yes/no or correct/incorrect data and estimates how different factors influence the likelihood of a correct judgment. The two factors tested were Technique (Real, Outpainting, Image-to-Video, Text-to-Video) and Theme (Film, History, Environment). We also looked at whether these factors interacted, for example, whether certain AI methods were harder to identify in specific themes. To control for multiple significance tests within the regression model, p-values were adjusted using the Holm–Bonferroni correction. We calculated average detection accuracy (mean ± standard deviation) for each combination of technique and theme. These results were visualized in a grouped bar chart (Figure 4) to show how recognition performance changed across conditions.

## 4.1 Overall detection accuracy

Participants showed moderate ability to distinguish AI-generated videos from real footage. Across all conditions, the mean accuracy was 0.60 (60%). Performance varied substantially by generation technique. Participants were best at recognizing motion-interpolated sequences and least confident when viewing text-generated content. Image-to-Video clips were identified most accurately ($M = 0.79$), followed by real clips ($M = 0.64$). Outpainting clips were identified correctly about half the time ($M = 0.49$) and Text-to-Video clips were the most difficult to classify ($M = 0.43$).

## 4.2 Influence of theme

The visual context strongly affected detection. In general, accuracy was higher for film and history scenes than for environment scenes. Real environmental footage was most often misjudged as AI (≈ 0.48 accuracy), suggesting that natural scenery may appear synthetic even when authentic. For AI videos, detection rates also depended on the theme: Outpainting was easiest to detect in environment clips (0.61) but dropped sharply in historical scenes (0.32). Image-to-Video remained consistently high across all themes (≈ 0.75–0.83), and Text-to-Video stayed low overall but improved slightly in film scenes (0.55). This suggests that thematic cues such as lighting, camera motion, and context realism shape how viewers interpret authenticity.

### 4.3 Influence of theme-technique interaction

The logistic regression confirmed significant effects of both technique and theme, as well as their interaction ($p < 0.001$). Compared to the reference condition (Real–Film), Image-to-Video clips had higher odds of being correctly identified (OR = 1.13, n.s.), while Text-to-Video (OR = 0.50, $p = 0.001$) and Outpainting (OR = 0.45, $p = 0.001$) had lower odds of correct detection. The environmental context significantly reduced accuracy overall (OR = 0.37, $p < 0.001$).

Several interactions were notable: Image-to-Video × Environment (OR = 8.22, $p < 0.001$) and Outpainting × Environment (OR = 6.89, $p < 0.001$) indicated that these AI methods became easier to recognize in natural scenes. Text-to-Video × History (OR = 0.44, $p = 0.003$) and Outpainting × History (OR = 0.41, $p < 0.001$) showed that both were harder to detect in historical contexts. All reported p-values were adjusted for multiple comparisons using the Holm–Bonferroni correction to maintain statistical reliability.

Overall, participants tended to underestimate real footage and overestimate AI involvement, indicating a general bias toward assuming synthetic origins. Yet detection accuracy clearly depended on how the video was produced and what it depicted. Image-to-Video sequences contained more recognizable motion cues that supported identification, whereas text-to-video and outpainting models produced content that blurred perceptual boundaries between real and artificial imagery. These results indicate that perceived realism in AI video is context-dependent rather than model-dependent and certain themes conceal AI artifacts more effectively than others.

## 5 Discussion

Participants frequently labelled real videos as AI-generated, achieving only 64% accuracy for authentic clips. This confirms an over-attribution bias, a tendency to doubt authenticity when participants know that AI content may be present. Similar “frequency-illusion” effects have been reported in studies of misinformation and deepfake detection, where awareness of manipulation lowers trust in real content (Hussein & Özad., 2025; Wasi et al., 2025). Once viewers recognize that AI generation is possible, even authentic imagery can appear synthetic.

Among AI generation methods, image-to-video clips were identified most accurately, suggesting that their motion patterns or texture continuity contain perceptual clues distinct from camera-captured footage. Text-to-Video and Outpainting, in contrast, blurred the distinction between real and artificial imagery. Their lower detection rates might imply that viewers lacked reliable heuristics for judging such content, as both methods produce novel imagery unconstrained by an underlying real reference.

Detection performance also depended strongly on thematic context. Film and historical scenes have higher accuracy, while environmental scenes consistently reduce it. One reason could be that natural landscapes are often generated to pristine perfection in AI methods, which can be easily identified.

The strong Technique × Theme interaction indicates that generative artifacts are not equally visible across visual domains; what seems realistic in one context may appear artificial in another.

## 6 Conclusions

From a design perspective, these findings highlight that realism and authenticity are not fixed properties of an AI model but emergent perceptual outcomes shaped by context, expectation, and content type. For creative practitioners, this suggests that AI video tools can be integrated intentionally to evoke different levels of perceived realism in different situations. For educators and media-literacy initiatives, the study emphasizes the need to train audiences to recognize model-specific artifacts and contextual cues rather than relying on intuition alone.

The study focused on short, silent clips viewed online, which may limit ecological validity. Future work should include additional techniques such as AI-based inpainting, ControlNet, and depth-aware video, with more advanced tools such as Cora 2 to gain a more comprehensive understanding, as well as expand the range of themes examined to uncover broader perceptual patterns. Expanding participant demographics and including confidence ratings would also clarify how subjective certainty relates to objective accuracy. Finally, longitudinal studies could explore whether exposure or training improves detection performance across model types. Combining technical innovation with context-driven human studies ensures the responsible deployment of generative video and equips creators and audiences to navigate synthetic realities.

**About the Authors:**

**Danyal Sarfraz:** Master's researcher in Design at UNIST. His research focuses on spatial computing, integrating 3D design, XR, and AI-based tools to enhance design workflows. He is currently developing methods that connect design prototyping with engineering validation using augmented reality.

**Raja Mubashir Karim:** Professor at NUST and Head of Research at School of Art, Design, and Architecture (SADA). His research focuses on colour maps, data visualization and user experience, currently teaching studio and research courses as an instructor.

**KwanMyung Kim:** Full professor in the Department of Design at UNIST and director of the Integration and Innovation Design Lab. He is also the CEO of ID SPACE Corp. With his strong practical knowledge and experience, he focuses on the design of innovative products and services through multidisciplinary approaches among industrial design, ergonomics and engineering design.

**Acknowledgement:** This work was supported by the Technology Innovation Program(RS-2025-04752997, Development of Design-Production Process Collision Element Prediction and Optimization Solution Technology) funded By the Ministry of Trade, Industry & Energy(MOTIE, Korea)